\documentclass[runningheads]{llncs}
\usepackage{graphicx}

\usepackage{tikz}
\usepackage{subfigure}
\usepackage{comment}
\usepackage{amsmath,amssymb} 
\usepackage{color}
\usepackage[linesnumbered,ruled,vlined]{algorithm2e}
\usepackage{booktabs}
\usepackage{multirow}
\usepackage{xcolor,colortbl}
\usepackage{caption}
\usepackage{subfigure}

\usepackage[accsupp]{axessibility}  
\newcommand{\etal}{{et al}.\@ }

\begin{document}
\pagestyle{headings}
\mainmatter
\def\ECCVSubNumber{3573}  

\title{Federated Selective Aggregation for Knowledge Amalgamation} 

\titlerunning{FedSA}
\author{Donglin Xie\inst{1*} \and
Ruonan Yu\inst{1*} \and
Gongfan Fang\inst{1} \and
Jie Song\inst{1} \and
Zunlei Feng\inst{1} \and
Xinchao Wang\inst{2} \and
Li Sun\inst{1} \and
Mingli Song\inst{1\dag}}
\authorrunning{D. Xie, R. Yu et al.}
%
\institute{Zhejiang University \and 
National University of Singapore \\
\email{\{donglinxie, ruonan, fgf, sjie, zunleifeng, lsun, brooksong\}@zju.edu.cn}
\email{xinchao@nus.edu.sg}}
\maketitle
\let\thefootnote\relax\footnotetext{* Co-first Authors. Donglin Xie and Ruonan Yu contributed equally to this work.}
\let\thefootnote\relax\footnotetext{$\dag$ Corresponding Author.}
\begin{abstract}
    In this paper, we explore a new 
    knowledge-amalgamation problem,
    termed Federated Selective Aggregation (FedSA). 
    The goal of FedSA is to  
    train a student model for a new task with the help of 
    several decentralized teachers, whose pre-training tasks
    and data are different and agnostic.
    Our motivation for investigating such a problem setup stems 
    from a recent dilemma of model sharing.
    Many researchers or institutes
    have spent enormous resources
    on training large and competent networks.
    Due to the privacy, security, or intellectual property issues,
    they are, however, not able to share their
    own pre-trained models, even if they wish to
    contribute to the community.
    The proposed FedSA offers a solution to
    this dilemma and makes it one step further
    since, again, the learned student may specialize in a
    new task different from all of the teachers.
    To this end, we proposed a dedicated strategy
    for handling FedSA. Specifically, 
    our student-training process is driven by a novel
    saliency-based approach that adaptively 
    selects teachers as the participants and 
    integrates their representative capabilities into the student. 
    To evaluate the effectiveness of FedSA, 
    we conduct experiments on both single-task 
    and multi-task settings. Experimental results demonstrate that 
    FedSA effectively amalgamates knowledge
    from decentralized models and achieves 
    competitive performance to centralized baselines.
\end{abstract}

\section{Introduction}

The past few years have witnessed tremendous progress in deep learning in many if not all machine learning applications, such as computer vision~\cite{krizhevsky2012imagenet},  natural language process~\cite{devlin2018bert},  and speech recognition~\cite{amodei2016deep}. The success of deep learning, in reality, is significantly attributed to  the model-sharing convention of the community: adopting a well-behaved and publicly-verified network
pre-trained by others, one may significantly reduce the retraining effort and enjoy the favorable performance resulting from thousands of GPU hours in a plug-and-play manner.

Nevertheless, such an increasing trend of model-sharing confronts 
a dilemma. Many researchers or institutes
have spent enormous resources
on training powerful networks.
However, due to issues like privacy, security, or intellectual property,
they cannot share their models with the public,
even it is their interest to contribute to the community.
Existing model-reuse strategies, such as 
knowledge distillation~\cite{hinton2015distilling}, transfer learning~\cite{ahn2019variational}, and domain adaptation~\cite{wang2018deep},
typically require pre-trained models to be fully available
and thus are inapplicable in this case.

\begin{figure}
  \centering
  \includegraphics[width=0.75\linewidth]{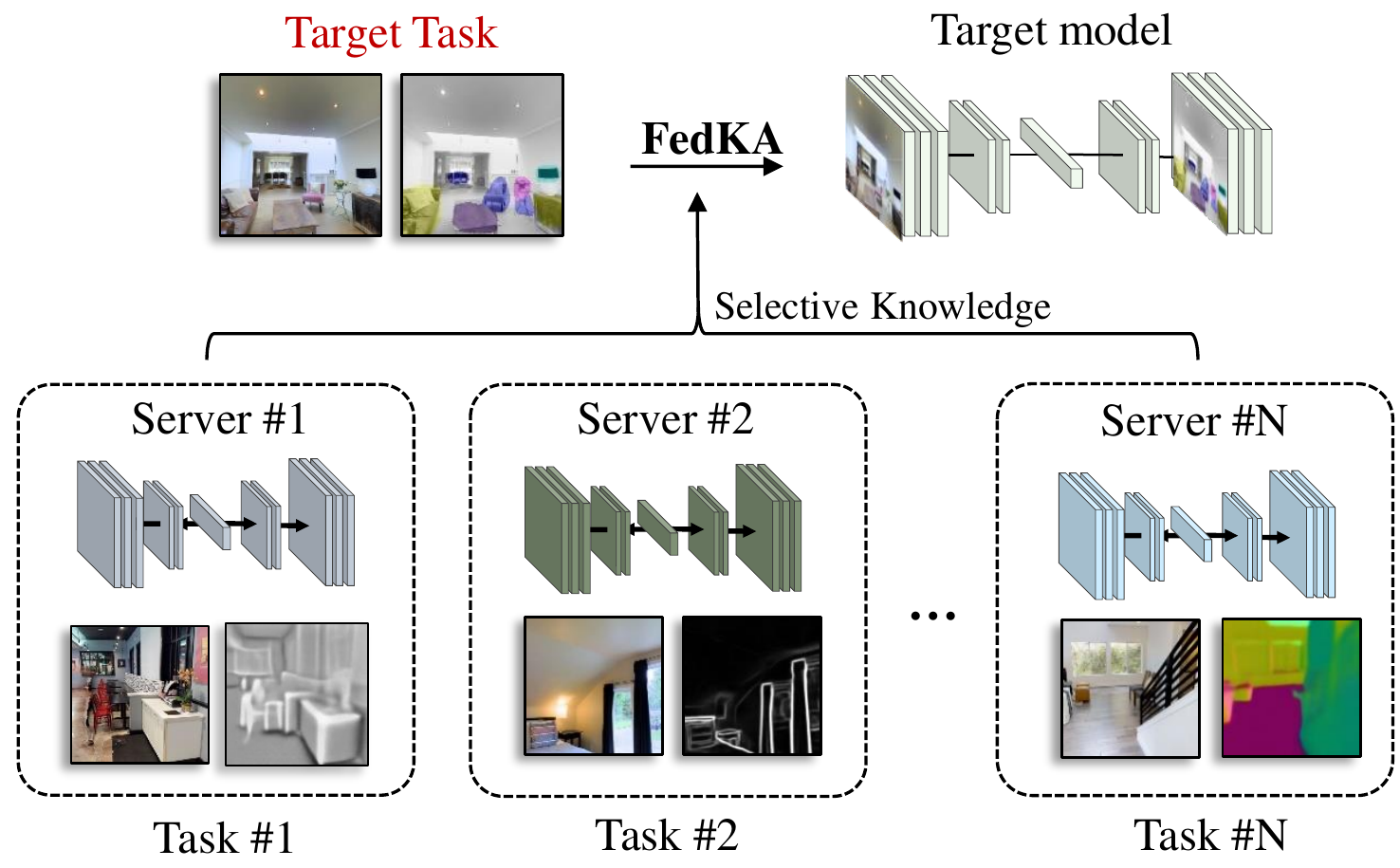}
  \caption{FedSA aims to train a student model with the help of selective knowledge from pre-trained but private models. The target task and training data can be different from private teachers, allowing flexible model customization in downstream tasks.}\label{fig:intro}
\end{figure}

To remedy this issue, we introduce a new knowledge-amalgamation task, termed Federated Selective Aggregation~(FedSA), whose aim is to
train a student model with the help of several decentralized teachers. 
We further relax the constraint on the specialization of the student: under FedSA, we allow the student to tackle a new task different from any of the teachers. To ensure that model information is not leaked during training, details of pre-trained models, including their pre-training tasks and data, should also be kept private and thus agnostic to other participants. Such a problem setup inevitably imposes enormous challenges to the federated model reusing due to difficulty finding helpful teachers for target tasks. Furthermore, as the training domain of different teachers diverges, the student training would have to account for balancing the knowledge aggregated from different teachers.

Notably, unlike conventional federated learning frameworks that only focus on data privacy, the proposed FedSA approach also takes model privacy into account. It enables model knowledge transfer without direct access, which provides a flexible and secure way for model sharing. 
To validate the effusiveness of our proposed approach, 
we conduct extensive experiments on single-task and multi-task datasets, including CIFAR, ImageNet, and Taskonomy. The results show that our approach successfully amalgamates proper knowledge from teachers to students. 

In short, our main contribution is a
dedicated FedSA approach for 
training students handing new tasks
from decentralized teachers.
Our approach does not require access to private models
or their origin training data and is able to utilize their knowledge to tackle downstream tasks. 
Experiments show the high scalability of FedSA in both 
single-task and multi-task settings.

\section{Related Work}

\paragraph{\textbf{Federated Learning.}} 
Federated learning ~\cite{mcmahan2017communication,kairouz2021advances,augenstein2020generative} develops a decentralized training schema for privacy-preserving learning, enabling multiple clients to learn a network collaboratively without sharing their private data. Note that conventional federated learning frameworks require the models of different clients to have the same architectures~\cite{li2019survey}, which hinders the model customization for different applications. Li \etal \cite{li2019fedmd} proposed FedMD to tackle the heterogeneous models through knowledge distillation. In some cases, data from different clients can be non-i.i.d, making conventional federated learning challenging to converge. Smith \etal \cite{smith2018federated} introduced federated multi-task learning and designed a robust and systems-aware optimization method to address this problem, termed MOCHA, to handle practical systems issues. Further, federated transfer learning~\cite{saha2021federated} is another topic related to this work, which is proposed to transfer knowledge across domains in federated settings. This work focuses on a more challenging setting, where both model-privacy and data-privacy are considered, and all information, including data domain, tasks, and model architectures, are agnostic.

\paragraph{\textbf{Knowledge Amalgamation.}} 
Knowledge amalgamation focuses on a student learning problem from multiple teachers models from different domains~\cite{ye2019student,thadajarassiri2021semi,li2021kabi,shen2019amalgamating,luo2019knowledge,jing2021amalgamating}. It aims to train a versatile student model that can handle all tasks from teachers. For example, Shen \etal \cite{shen2019amalgamating} proposed a layer-wise coding approach to fuse the representation of different tasks and learn a comprehensive classifier from multiple teacher classifiers. Luo \etal \cite{luo2019knowledge} introduced the idea of common feature learning to extract shared knowledge between teachers. Ye \etal \cite{ye2019student} proposed a multi-task amalgamation method, which learns a superior student model from teachers to handle various tasks, including semantic segmentation, depth estimation, and surface normal estimation. Knowledge amalgamation has recently been extended to non-euclidean data like the graph, where a multi-talent student model is trained for both point cloud classification and segmentation~\cite{jing2021amalgamating}. In this work, we study a new setting for knowledge amalgamation, where the teacher models are not directly available. 

\paragraph{\textbf{Model Reusing.}} 
As an increasing number of pre-trained models have been released by developers, reusing pre-trained models has become a common practice in the community for solving downstream tasks. The seminal work of Hinton~\cite{hinton2015distilling} established the basic framework of knowledge distillation, which aims to learn a lightweight student model from cumbersome teachers. Following this pioneering framework, many algorithms have been proposed for model reusing. In particular, Ahn \etal \cite{ahn2019variational} proposed VKD to transfer the representative ability of a pre-trained model to a student. Lopes \etal \cite{lopes2017datafree}  proposed data-free knowledge distillation to reuse public models without accessing the original training data. Jiao \etal \cite{jiao2020tinybert} utilized knowledge distillation to compress a transformer model for efficient inference. This work considers a selective model reusing problems in federated settings and proposes FedSA to train a student model on a different task.

\begin{figure*}[t]
  \centering
  \includegraphics[width=\linewidth]{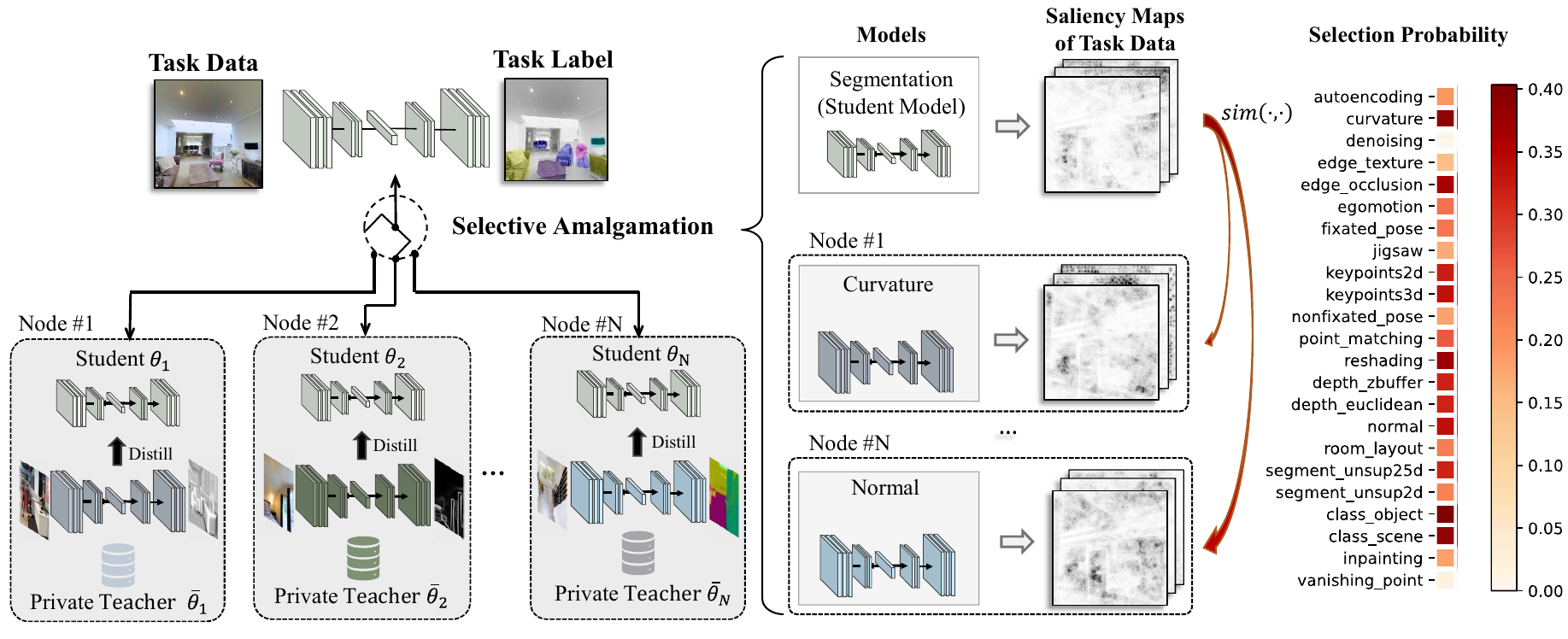}
  \caption{Overview of the FedSA framework.
  It comprises three parts: knowledge selection, selective aggregation, and target task adaptation. Knowledge selection picks proper models with saliency maps,
  selective aggregation distills the knowledge from the pre-trained models to local students,
  while task adaptation strengthens the performance of the target model.}
  \label{fig:framework}
\end{figure*}

\section{Methodology}

\subsection{Problem Setup}
Let $\boldsymbol{\mathcal{M}}=$ $\{M_1,M_2$ $,...,M_N\}$ denote 
a set of model candidates on distinct tasks.
The models are pre-trained on the datasets $\boldsymbol{\mathcal{D}}= \{D_1,D_2,...,D_N\}$. 
The models and datasets can only be visited on their server. The goal is to obtain a model $M_\mathcal{T}$ that excels on the target task $\mathcal{T}$ while protecting the model and data privacy.
For ease of description, let 
$\boldsymbol{\overline{\theta}_n}$
indicates the parameters of $n$-th pre-trained model, and $\boldsymbol{\theta_\mathcal{T}}$ indicates the parameters of the target model. The annotated data of target task $\mathcal{T}$ is  $D_\mathcal{T}=\{\{\boldsymbol{x_1},y_1\},\{\boldsymbol{x_2},y_2\},...,\{\boldsymbol{x_M},y_M\}\}$. The expensive labeling costs cause the amount of annotated data to be usually limited. 

\subsection{Federated Selective Aggregation for Knowledge Amalgamation}
\paragraph{\textbf{Overview.}} 
We propose a novel framework named Federated Selective Aggregation~(FedSA).
It utilizes the knowledge of pre-trained models to train a target model
while protecting the model and data privacy. 
An overall framework of FedSA is shown in Figure \ref{fig:framework}. 
Because of the significant gaps between different tasks, massive variation in knowledge exists among networks. 
For the target model, the effects of knowledge from different models are vastly different. The knowledge from a proper model can be better generalized to the target task. Therefore, finding a way to select proper knowledge for the target task is of great importance. 

Due to privacy issues, models and data may only be allowed for internal usage. Without access to the model and data, knowledge amalgamation methods cannot be directly deployed, inevitably making the aggregation of knowledge from the selected pre-trained models more challenging. The selective aggregation uses a local student as the messenger of communication, and only the local student is exchanged so that the knowledge aggregation is entirely local and private. Besides, the collaboration of different pre-trained models further guarantees the model's privacy. However, although proper knowledge has been chosen and aggregated, the gap between the target and local tasks still exists. Hence, it is necessary to adapt the target model to the target task, which significantly improves the performance of the target model.

\paragraph{\textbf{Knowledge Selection.}}
Knowledge selection is to select proper pre-trained models for the target task. To select proper pre-trained models, the target model should learn the knowledge about the target task first. In detail, the target model is trained with limited annotated target data before the knowledge selection. Then the saliency map is used to choose proper models. Saliency map\cite{song2019deep} is widely used to measure the knowledge transferability between heterogeneous neural networks. The saliency map is based on the attribution methods\cite{selvaraju2017grad,chattopadhay2018grad}, which assign importance scores to the input for a specific output. The transferability score between different models can be calculated for a specific probe data set. The suitability of model knowledge is positively correlated with the transferability score. Owing to limited target data, the knowledge selection is inaccurate at first. Therefore, the knowledge selection is performed dynamically.

Let $a_{n,j}^k \in \mathbb{R}^{WHC}$ indicates saliency map for one input image $\boldsymbol{x_j}$ in the probe data $\mathcal{D}_\mathcal{T}$. It can be computed through one single forward and backward propagation,
\begin{equation}
    a^k_{n,j}=[\lvert{ \frac{\partial r_k}{\partial \boldsymbol{x_j}} }\rvert]^{W_n H_n C_n},
    \label{eq:smap}
\end{equation}
where $r_k$ is the network's hidden representation of the $k$-th layer. 
Let $A_n^k$ denote the overall saliency map for the $k$-th layer of $n$-th model $M_n$. It can be formed by averaging the all saliency maps $a_{n,j}^k|_{j=1}^M$. Formally, we have $A_n^k=\frac{1}{M}\sum_{j=1}^M{a_{n,j}^k}$. 
After measuring $N$ attribution maps for the models in $\mathcal{M}$, we get  $\mathcal{A}=\{A_1^k,A_2^k,...,A_N^k\}$. The distance of two models can be estimated as follows: 
\begin{equation}
    d(M_n^k,M_\mathcal{T}^k)=\frac{M}{cos\_sim(A_n^k,A_{\mathcal{T}}^k)},
    \label{eq:m_distance}
\end{equation}
where $cos\_sim(A_{n}^k,A_{\mathcal{T}}^k)=\frac{A_{n}^k\cdot A_{\mathcal{T}}^k}{||A_{n}^k||\cdot||A_{\mathcal{T}}^k||}$, $M$ is the number of target data. The pairwise transferability can be derived based on these distances. For the model-wise knowledge selection, $K$ models are chosen from the model candidate set $\boldsymbol{\mathcal{M}}$. To select models with great transferability, the selected model indexes set $\hat{\mathcal{S}}$ is established by replacement sampling with probabilities $\{p_1,p_2,...,p_N\}$. The probabilities is the normalized transferability score between the target model $M_\mathcal{T}$ and candidate models $\{M_n\}_{n=1}^N$. The sampling probabilities are formed as:
\begin{equation}
    p_n=\frac{e^{Q_{n,\mathcal{T}}}}{\sum_{b=1}^Ne^{Q_{b,\mathcal{T}}}},
\end{equation}
where $Q_{n,\mathcal{T}}$ means the transferability score between $n$-th model and target model. It is formed as  $Q_{n,\mathcal{T}}=\frac{1}{d(M_n^k,M_\mathcal{T}^k)}$. The theoretical analysis in the supplemental materials indicates that this replacement sample policy guarantees the convergence of the algorithm.

\paragraph{\textbf{Selective Aggregation.}}
After the knowledge selection, the selected model set $\hat{\mathcal{S}}$ has been established. In selective aggregation, the knowledge is distilled from the selected models to the target model while protecting the model and data privacy. A neural network called the local student model is employed as the substitute for the target model. It behaves as the messenger of communication between the target and pre-trained models. The local student model learns the knowledge from the pre-trained models and is uploaded to the central server for further  
aggregation. The local student model avoids direct access, 
ensuring the model and data privacy.  

The architecture of most neural networks can be divided into the encoder and the decoder. The encoder extracts the abstract features of input images, and the decoder maps the abstract features to the task-related results. The output features of the encoder include rich and semantic information. This feature extraction ability is general in most cases. Therefore, the  FedSA lets the local student learn the knowledge from the local teacher model through the feature-based knowledge distillation. The feature-based knowledge distillation \cite{romero2014fitnets,park2020feature} uses the output of the intermediate layers as the supervision. In detail, the intermediate features of the teacher model provide hints\cite{romero2014fitnets} and the intermediate features of local students are forced to be matched with teachers' hints. 
Through feature-based knowledge distillation, the feature extraction ability of the teacher model can be distilled to the student model. 

Because of the heterogeneous network architectures, the output feature dimensions between the teacher and student models could be distinct. In order to align the output dimensions of the teacher and that of the student, a translator block is used. The translator block
is formed by three convolutions with $1 \times 1$ kernel. Through the block, the student output is converted to a predefined output length. The $\mathbb{F}_n^s$ and $\mathbb{F}_n^p$ indicate the aligned features of the student model and teacher model $M_n$ respectively. The $\boldsymbol{X}$ denotes the training data of model $M_n$. The  $\boldsymbol{\theta_n^t}$ indicate the parameters of the $n$-th local student in $t$-th iteration. To encourage the intermediate output of the student model to imitate that of the teacher model, the loss of feature knowledge distillation is computed as follows:
\begin{equation}
    \mathcal{L}_{KD} =\frac{1}{2}||\mathbb{F}_n^p(\boldsymbol{X};\boldsymbol{\overline{\theta}_n}) - \mathbb{F}_n^s(\boldsymbol{X};\boldsymbol{\theta_n^t})||^2 .
    \label{eq:KD_loss}
\end{equation}
After the local knowledge distillation, the parameters of local student $\boldsymbol{\theta_n^t}$ are uploaded to the center server.
To aggregate the knowledge of different local student models, the update of the target model's parameters with $K$ selected teachers is as follows:
\begin{equation}
    \boldsymbol{\theta_{t+1}}=\frac{1}{K}\sum_{n\in\hat{\mathcal{S}}}\boldsymbol{\theta_n^t},
    \label{eq:avg}
\end{equation}

\SetKwInput{KwInput}{Input}                
\SetKwInput{KwOutput}{Output}              
\begin{algorithm}[t]
\DontPrintSemicolon
  
  \KwInput{number of communication rounds $T$, number of local epoch $E$, selective gap $E_g$, local learning rate $\eta_T$, center learning rate $\eta_c$, number of selected model $K$}
  \KwOutput{Target model $\boldsymbol{\theta_\mathcal{T}}$}
  \KwData{Target dataset $\mathcal{D}_\mathcal{T}$}

  \textbf{Center server executes:}
  
  \For{$t=0,1,2,...,T-1$}
  {
    \If{$t \% E_g == 0$}
    {
        Train weak target model $\boldsymbol{\theta_t}$ on target dataset $\mathcal{D}_\mathcal{T}$ 
        
        Generate selected indexes set $\hat{\mathcal{S}}$ with saliency map
    }  
    \For{$n \in \hat{\mathcal{S}}$ \textbf{in parallel}}
    {
        Send the global model $\boldsymbol{\theta_t}$ to $n$-th server
        
        $\boldsymbol{\theta_n^t}\xleftarrow{}\textbf{SelectiveKA}(n,\boldsymbol{\theta_t}) $ 
    }
    $\boldsymbol{\theta_{t+1}}\xleftarrow{}\frac{1}{K}\sum_{n\in\hat{\mathcal{S}}}\boldsymbol{\theta_n^t}$
  }
  Target task adaptation with $\mathcal{D}_\mathcal{T}$
  
  return parameters of target model $\boldsymbol{\theta_\mathcal{T}}$
  
  $\textbf{SelectiveKA}(n,\boldsymbol{\theta_t}):$
  
  $\boldsymbol{\theta_n^t} \xleftarrow{}\boldsymbol{\theta_t}$
  
  \For{epoch $e=1,2,...,E$}
  {
    
    \For{each batch $\boldsymbol{b}=\{\boldsymbol{x},y\}$ of $D_n$}
    {
        $\mathcal{L}_{KD}\xleftarrow{}\frac{1}{2}||\mathbb{F}_n^p(\boldsymbol{x};\boldsymbol{\overline{\theta}_n}) - \mathbb{F}_n^s(\boldsymbol{x};\boldsymbol{\theta_n^t})||^2$
    
        $\boldsymbol{\theta_n^t}\xleftarrow{}\boldsymbol{\theta_n^t}-\eta_t\nabla\mathcal{L}_{KD}$
    }
  } 
  return  $\boldsymbol{\theta_n^t}$ to center server

\caption{The FedSA framework}
\end{algorithm}

\paragraph{\textbf{Task Adaptation.}}
In the selective aggregation, the target model $M_\mathcal{T}$ only learns the knowledge about generalizable feature extraction. So it is natural to make target task adaptation for more incredible performance. In detail, the decoder part of the target model is trained on the annotated target data. In this stage, the target model learns to map features to the target labels. 

\paragraph{\textbf{Algorithm Summary.}}
In the framework of FedSA, the target model is trained with the limited labeled target data at first. Then the proper models are chosen for further selective aggregation. Owing to the progressive learning of the target model, the process of knowledge selection and selective aggregation is iterative. It is worth noting that the selected knowledge is dynamic. After several rounds of selective aggregation, the target model is adapted to the target data.

\section{Experiments}
In this section, we validate the proposed method FedSA and evaluate the performance via two sets of quantitative experiments, single-task amalgamation, and multi-task amalgamation. Single-task amalgamation experiments focus on classification tasks, where pre-trained local private teacher models and the target student model are responsible for different classes. Moreover, the target student model is expected to solve a new visual task, also called the target task, in multi-task amalgamation experiments. The target task is different from the tasks of pre-trained local private teachers. Further, we also conduct some ablation, analytical, and visualization experiments to validate the effectiveness of FedSA. More experimental details can be found in our supplementary material.

\subsection{Experimental Settings}

\paragraph{\textbf{Datasets and Models.}}
The single-task amalgamation experiments are conducted using three public benchmark datasets, i.e.,  CIFAR10, CIFAR100, and ImageNet. We adopt ResNet34\cite{he2016deep} as the network structure of local private teachers, and those teachers will be pre-trained on ten non-overlapped subsets of ImageNet. ResNet34\cite{he2016deep}, ResNet18\cite{he2016deep}, VGG19\cite{simonyan2014very}, WRN\_16\_2\cite{zagoruyko2016wide}, and MobileNetV2\cite{sandler2018mobilenetv2} will be the network structures of local student models and target model. Specifically, Each subset of ImageNet contains $32\times 32$ images from 100 classes of ImageNet. In our experiments, CIFAR10 and CIFAR100 are used as target datasets as well as the probe datasets, and all ImageNet subsets are considered as non-IID private sets, which can not be accessed by other nodes. 
Experiments for multi-task amalgamation are performed on Taskonomy datasets\cite{zamir2018taskonomy} in a similar protocol. Taskonomy datasets include over 4 million images of indoor scenes from around 600 buildings, and each image has 26 annotations for all kinds of visual tasks. We use 25 off-the-shelf ResNet50 with different decoders as teacher models and train a student for a new task.

\paragraph{\textbf{Implementation Details.}}
All the methods are implemented in PyTorch on a Quadro P6000 GPU in the experiment. The target task in single-task amalgamation is classification, and the experiments are conducted on both homogeneous and heterogeneous networks. A group of models in ResNet34 network architecture is pre-trained by 100 classes of data from the ImageNet dataset. The ImageNet dataset is sorted based on labels and then distributed to the models on a class-by-class basis. In the homogeneous network experiments, local student models have the same network architecture as the local teacher models, while in the heterogeneous network experiments, ResNet18, VGG19, WRN\_16\_2, and MobileNetV2 are used for the network architectures of the local student models. As for multi-task amalgamation, a task from the task dictionary is chosen as the target task, and the pre-trained models of the remaining tasks are used as the local teacher models. 

\subsection{Single-Task Amalgamation}

\begin{table*}[t]
\renewcommand{\arraystretch}{0.9}
\centering
\resizebox{\textwidth}{!}{
\begin{tabular}{c l r r r r r r r r r}
\toprule
\multirow{2}{*}{\bf Dataset} & \multirow{2}{*}{\bf Method} & \multicolumn{4}{c}{\bf Homogeneous} & & \multicolumn{4}{c}{\bf Heterogeneous (10\%)} \\
\cmidrule{3-6} \cmidrule{8-11}
\bf  &  & \bf 1\%  & \bf 5\% & \bf 10\%  & \bf 100\%  & & \bf Wrn & \bf Res & \bf Vgg  & \bf Mob \\
\midrule
\multirow{10}{*}{\rotatebox[origin=c]{90}{\bf CIFAR-10}} 
& Scratch Training  & 40.00  & 62.14  & 75.43  & 91.56 && 73.97 &  74.91 & 73.63 & 66.23 \\
& Transfer  & 63.70  & 79.38  & 83.34  & 92.88 && - & - & - & - \\
& Pseudo Label\cite{lee2013pseudo}  & 35.34  & 64.81  & 74.17  & - && 75.79 &  76.14 & 74.86 & 56.35 \\
& Noisy Student\cite{xie2020self}  & 11.05  & 43.46  & 61.95  & -   && 55.47 &  59.60 & 58.58 & 50.74\\
& FixMatch\cite{sohn2020fixmatch} & 40.96 & 58.59 & 72.16 & - && 80.72 & 79.39 & 74.06 & 68.89 \\
& Meta Pseudo Label\cite{pham2021meta} & 43.03 & 51.33 & 53.73 & - && 42.00 & 50.58 & 59.23 & 48.64 \\
&FedAvg$+$\cite{mcmahan2017communication} & 69.96 & 71.58 & 78.04 & 89.24 && 76.05 & 71.69 & 80.97 & 81.87 \\
&FedMD$+$\cite{li2019fedmd} & 71.25 & 83.20 & 85.52 & 92.37 && 81.28 & 84.23 & 80.76 & 71.17\\
&FedProx$+$\cite{li2020federated} & 36.51 & 55.88 & 63.05 & 87.07 && 46.11 & 65.11 & 61.17 & 49.47 \\

& \bf FedSA & \bf 77.15  & \bf 85.30  & \bf 87.57  &  \bf 93.83 & & \bf 84.70 & \bf 86.39  & \bf 85.71 & \bf 82.04 \\
\midrule
\multirow{10}{*}{\rotatebox[origin=c]{90}{\bf CIFAR-100}} 
& Scratch Training & 9.24  & 21.00  & 29.25  & 66.75 && 28.99 & 29.82 & 19.78 & 26.05 \\
& Transfer & 18.43  & 42.54  & 48.45  & 69.27 && - & - & - & - \\
& Pseudo Label\cite{lee2013pseudo} & 5.95  & 17.78  & 30.93  & - && 34.73 & 31.38 & 19.03 & 22.10 \\
& Noisy Student\cite{xie2020self} & 6.76  & 10.49  & 16.44  & -  && 16.69 & 15.36 & 14.85 & 14.51 \\
& FixMatch\cite{sohn2020fixmatch} & 8.93 & 20.34 & 28.80 & - && 35.71 & 31.23 & 19.35 & 12.91\\
& Meta Pseudo Label\cite{pham2021meta} & 6.84 & 13.91 & 15.78 & - && 7.76 & 13.96 & 11.96 & 13.23 \\
& FedAvg$+$\cite{mcmahan2017communication} & 20.93 & 29.35 & 36.08 & 62.91 && 42.76 & 34.46 & 49.85 & 44.32 \\
& FedMD$+$\cite{li2019fedmd} & 28.67 & 44.73 & 50.31 & 70.28 && 42.30 & 49.09 & 42.38 & 30.35\\
& FedProx$+$\cite{li2020federated} & 4.82 & 15.72 & 21.86 & 52.67 && 16.29 & 22.36 & 11.46 & 13.88 \\
& \bf FedSA & \bf 32.99  & \bf 52.76  & \bf 57.87 & \bf 73.49 &&  \bf 50.49 & \bf 58.32  & \bf 53.29 & \bf 47.25  \\
\bottomrule
\end{tabular}
}
\caption{
The performance of FedSA and baseline methods on CIFAR10 and CIFAR100. Symbol $+$ indicates that the method is modified for our problem settings. 
}
\label{tab:main}
\end{table*}

As for single-task amalgamation, the classification task is selected as the target task. Different local tasks refer to non-overlap classification classes, and the classification ability of local teacher models is transferred to help classify the new categories in the global target task. To show the effectiveness of the proposed method, we conduct the following experiments to compare with FedSA. 
The results are shown in Table \ref{tab:main}.

\paragraph{\textbf{Baselines.}}
Nine methods Scratch Training, Transfer, Pseudo Label\cite{lee2013pseudo}, Noisy Student\cite{xie2020self}, FixMatch\cite{sohn2020fixmatch}, Meta Pseudo Labels\cite{pham2021meta}, FedAvg$+$\cite{mcmahan2017communication}, FedMD$+$\cite{li2019fedmd} and FedProx$+$\cite{li2020federated} are adopted as baselines in experiments to validate the effectiveness of the proposed method.
For Scratch Training, we take \{1\%, 5\%, 10\%, 100\%\} of images from each class of CIFAR10 to form the probe dataset to train the models with ResNet34\cite{he2016deep} architecture directly, and so does with CIFAR100. 
The Transfer method takes the same method as Scratch Training to form the probe dataset but uses the obtained probe dataset to train the models pre-trained with 100 classes of samples from the ImageNet dataset.
Moreover, four semi-supervised learning methods, Pseudo Label\cite{lee2013pseudo}, Noisy Student\cite{xie2020self}, FixMatch\cite{sohn2020fixmatch} and Meta Pseudo Label\cite{pham2021meta} are adopted to demonstrate the competitive performance of FedSA with limited labeled data.
As shown in Table \ref{tab:main}, we take \{1\%, 5\%, 10\%\} of the dataset as labeled data and leave other samples as unlabeled ones.
As there is no method that directly addresses the problem, three federated methods, FedAvg$+$\cite{mcmahan2017communication}, FedMD$+$\cite{li2019fedmd} and FedProx$+$\cite{li2020federated} are modified for fair comparison. In order to keep the same privacy implication, the local update of FedAvg$+$ and FedProx$+$ is implemented with knowledge distillation. After the iterative federated learning phrase, the model from three federated methods is further fine-tuned on the downstream target data.  

Table \ref{tab:main} shows the results of the proposed method and baselines. The proposed FedSA outperforms other baselines on both homogeneous and heterogeneous network settings. As Table \ref{tab:main} shows, with only limited labeled data, the semi-supervised methods, including Pseudo Label, Noisy Student, FixMatch, and Meta Pseudo Label, have poor performance on all experiment settings. With the effectiveness of collaboration, the performance of FedAvg$+$ and FedMD$+$ is close to the proposed FedSA. The FedPorx$+$ is unable to achieve good performance on most experiment settings owing to the mismatch of the loss function and unfixed optimization steps.

\begin{table*}[t]
\renewcommand{\arraystretch}{0.9}
\centering
\resizebox{\textwidth}{!}{
\begin{tabular}{c l r r r r r r r r r}
\toprule
\multirow{2}{*}{\bf Dataset} & \multirow{2}{*}{\bf Strategy} & \multicolumn{4}{c}{\bf Homogeneous} & & \multicolumn{4}{c}{\bf Heterogeneous (10\%)} \\
\cmidrule{3-6} \cmidrule{8-11}
\bf  &  & \bf 1\%  & \bf 5\% & \bf 10\%  & \bf 100\%  & & \bf Wrn & \bf Res & \bf Vgg  & \bf Mob \\
\midrule
\multirow{7}{*}{\rotatebox[origin=c]{90}{\bf CIFAR-10}}
& Fixed+Unweighted & 73.46  & 82.22  & 85.28  & 93.11  && 83.89 & 84.00  & 83.78 & 80.49  \\
& Top-K+Unweighted  & 75.14  & 84.15  & 85.55  & 93.28 && 83.23& 85.76  & 84.07 & 81.76 \\
& Top-K+Positive Weighted  & 74.65  & 79.89  & 87.38  & 93.16 && 83.69 & 84.70  & 83.55 & 81.19 \\
& Top-K+Negative Weighted  & 76.54  & 83.52  & 85.87  & 93.51 && 83.36 & 85.29  & 84.26 & 81.50 \\
& Least-K+Positive Weighted & 75.50  & 84.44  & 85.16  & 93.21 && 82.78 & 85.72  & 83.85 & 81.14 \\
& Random+Positive Weighted  & 75.81 & 84.15 & 86.45 & 93.65 && 84.17 & 86.10  & 85.23 & 81.30 \\
& \bf FedSA & \bf 77.15  & \bf 85.30  & \bf 87.57  &  \bf 93.83 && \bf 84.70 & \bf 86.39  & \bf 85.71 & \bf 82.04 \\
\midrule
\multirow{7}{*}{\rotatebox[origin=c]{90}{\bf CIFAR-100}}
& Fixed+Unweighted& 28.42  & 48.72  & 54.05  & 71.38 && 49.08 & 54.97  & 48.71 & 44.21  \\
& Top-K+Unweighted & 32.78  & 49.10  & 55.01  & 71.99 && 50.37 & 55.18  & 50.04 & 44.33 \\
& Top-K+Positive Weighted & 32.80  & 51.22  & 54.50  & 71.67 && 50.24 & 54.42  & 49.88 & 45.37  \\
& Top-K+Negative Weighted & 32.69  & 50.22  & 56.13  & 72.44 && 49.39 & 55.34  & 51.16 & 45.24  \\
& Least-K+Positive Weighted & 32.74  & 50.95  & 55.61  & 71.85 && 49.76 & 55.74  & 49.57 & 45.88 \\
& Random+Positive Weighted &  32.46  & 52.39  & 56.77  & 72.97 && 50.03 & 57.76  & 51.98 & 46.96  \\
& \bf FedSA & \bf 32.99  & \bf 52.76  & \bf 57.87 & \bf 73.49 && \bf 50.49 & \bf 58.32  & \bf 53.29 & \bf 47.25  \\
\bottomrule
\end{tabular}
}
\caption{The performance of different amalgamation strategies. Accuracy of student models amalgamated from 10 homogeneous and heterogeneous ImageNet teachers.}
\label{tab:amal_strategy}
\end{table*}

\paragraph{\textbf{Ablation study on Amalgamation Strategy.}}
In order to show the effectiveness of the selection strategy, several methods with different knowledge selection and amalgamation are implemented for comparison. Apart from the proposed method FedSA, four selection strategies are considered,  
\emph{Random},
\emph{Fixed}, \emph{Top-K} and \emph{Least-K}. Three knowledge aggregation ways, \emph{Positive Weighted}, \emph{Negative Weighted} and \emph{Unweighted}, are implemented in this section. 
\emph{Random} means model selection with equal probability. 
\emph{Fixed} means the selected models are immutable. \emph{Top-K} means the selection of the k models with the highest similarity, while \emph{Least-K} with the lowest similarity. The similarities can also be seemed as influencing factors to calculate the weights for model aggregation. According to similarity, we have three different aggregation ways: weighted with positive correlation to similarity (referred to as \emph{Positive Weighted}), weighted with negative correlation to similarity (referred to as \emph{Negative Weighted}), and \emph{Unweighted}. 

Table \ref{tab:amal_strategy} shows the results of the methods with the different selection and amalgamation strategies. Overall, the performance of FedSA is quite competitive compared with other methods. For all settings, FedSA has the most outstanding performance. 
As Table \ref{tab:amal_strategy} shows, 
\emph{Fixed} 
gets the worst results due to the overfitting problem. 
\emph{Random} sometimes can achieve competitive results but is not applicable for practical applications with many pre-trained models.
\emph{Top-K}, \emph{Least-K} and the knowledge selection method adopted in FedSA select models according to similarities calculated by saliency maps. From the results, the models with high similarities in saliency maps are more advantageous for knowledge aggregation. Thus \emph{Least-K} knowledge chosen method is less competitive. However, as the global target model is updated, the selected models tend to be fixed with \emph{Positive Weighted}, which is detrimental to improving the effectiveness of the global model. As a result, FedSA, with stability and mobility, can select the models with high similarity and avoid the problem of over-fitting to some degree.
To further verify the effectiveness of the proposed method, extensive experiments on heterogeneous networks are conducted. 
Different nodes can have diverged model architectures. 
The heterogeneous network setup further completes the conception of network structure protection. 
From the results, the proposed method FedSA has the advantages of high accuracy and high stability. 

\begin{figure}
    \centering
    \includegraphics[width=\textwidth]{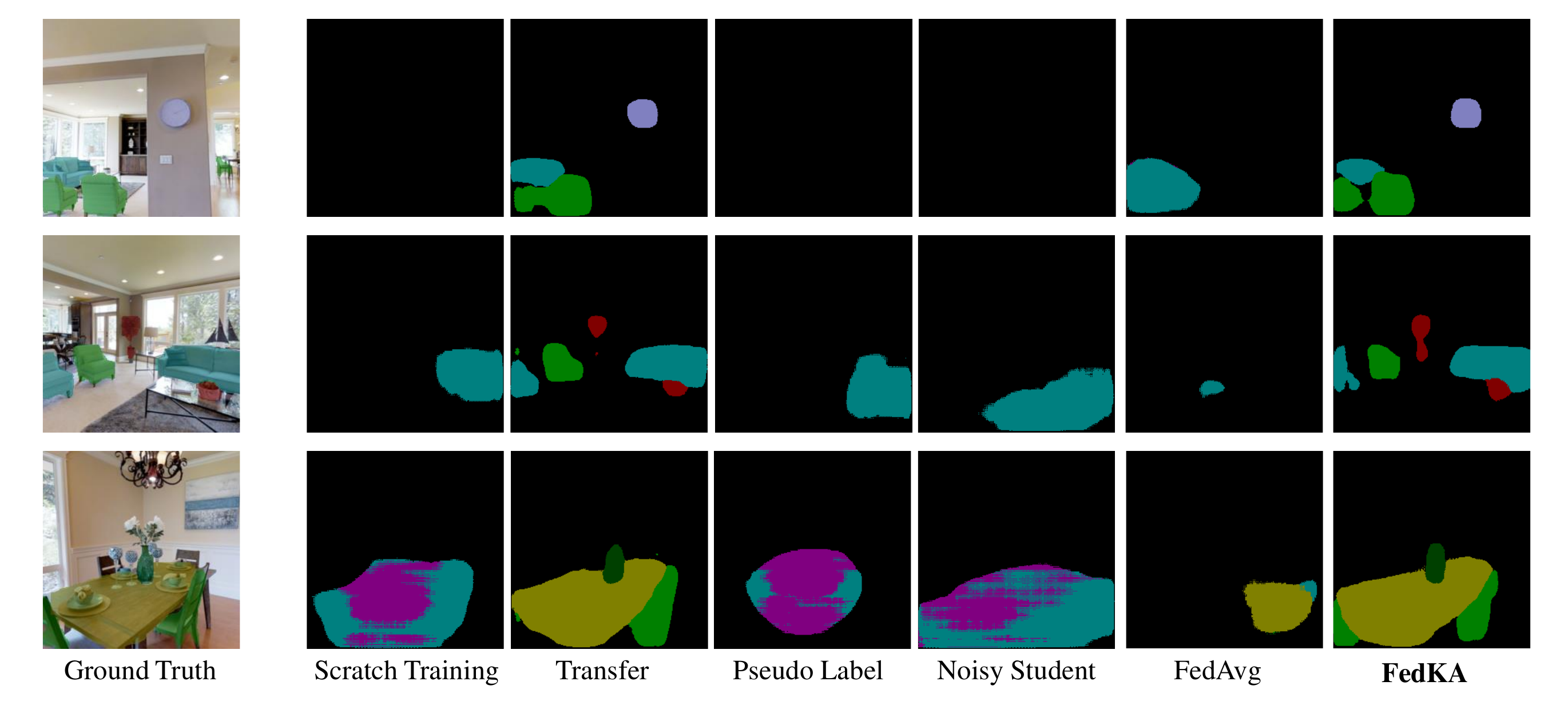}
    \caption{The visualization semantic segmentation results of proposed FedSA and some baselines. The transfer method requires the pre-trained models to be accessible.}
    \label{fig:vis_seg}
\end{figure}

\begin{figure}[t]
  \centering
  \subfigure[]{
  \includegraphics[width=0.35\linewidth]{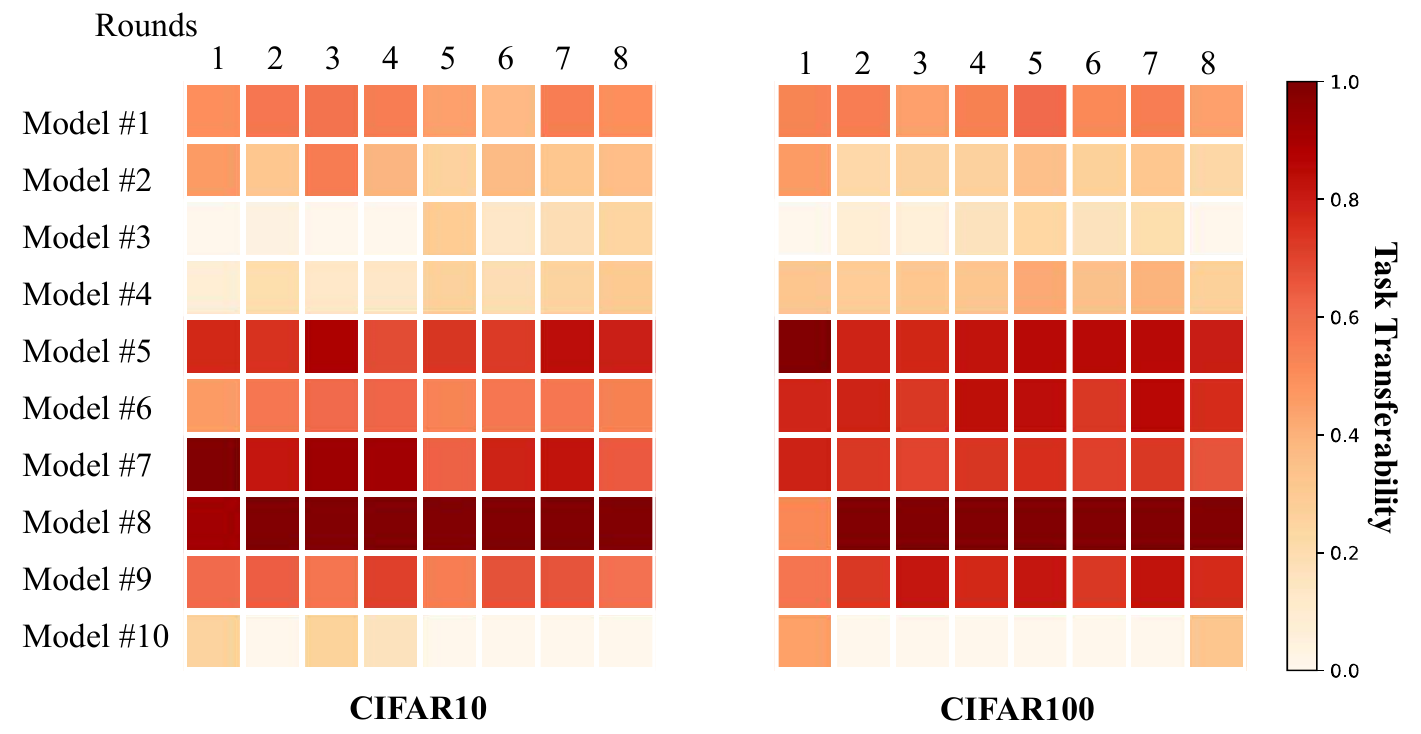}
  }
  \subfigure[]{
  \includegraphics[width=0.29\linewidth]{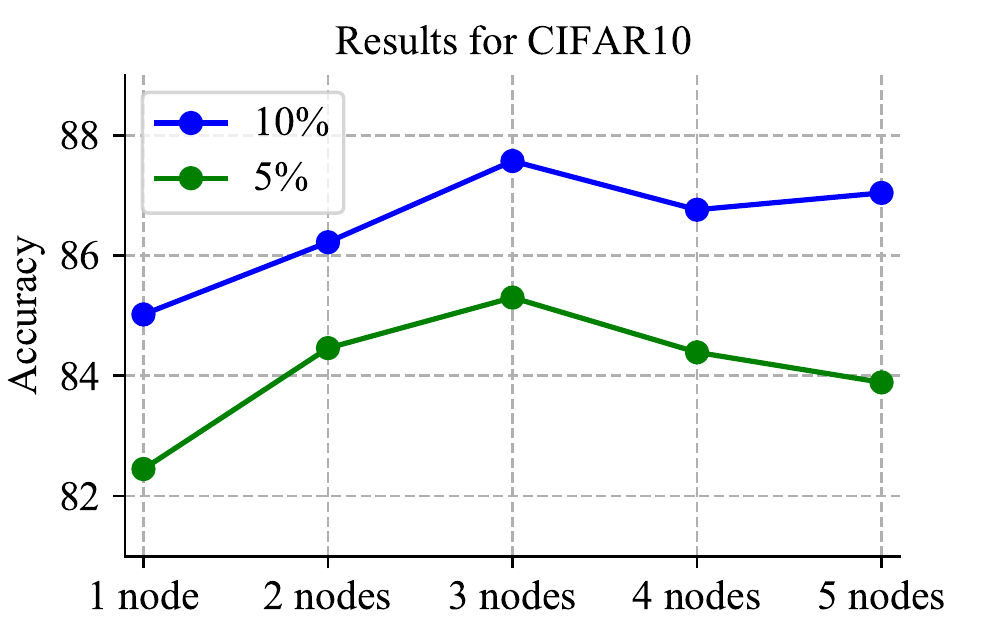} 
  }
  \subfigure[]{
  \includegraphics[width=0.29\linewidth]{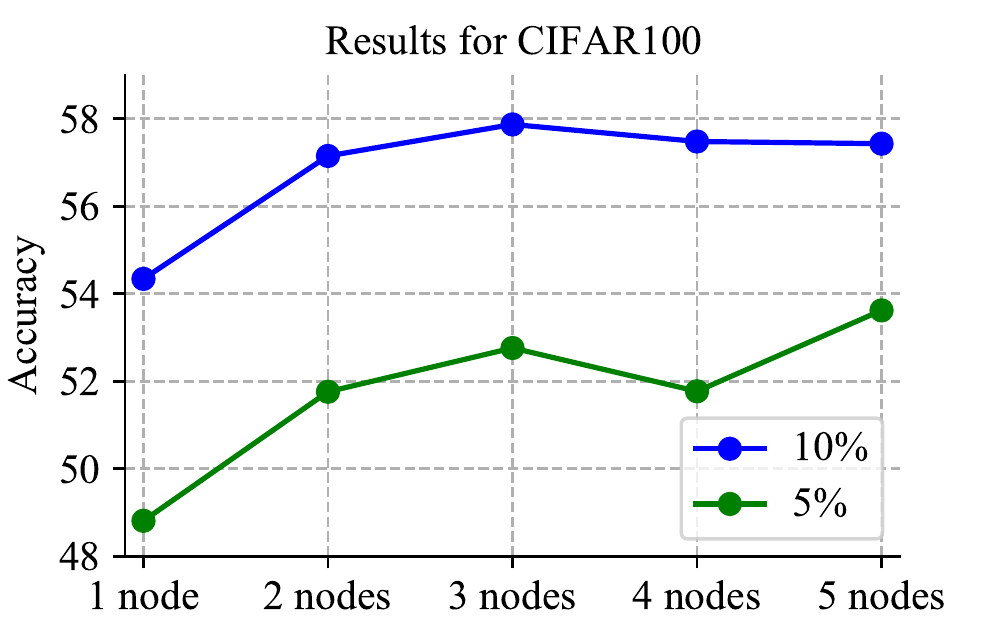} 
  }
  \caption{(a) Visualization of Dynamic Knowledge Selection for CIFAR10 and CIFAR100. The darker color indicates the higher transferability scores. (b)\&(c) The performance of FedSA with different number of nodes selected.}
  \label{fig:vis_trans}
\end{figure}

\begin{table*}[t]
\renewcommand{\arraystretch}{0.9}
\centering
\resizebox{\textwidth}{!}{
\begin{tabular}{c c | l c c c c c c c c c c }
\toprule
Dataset & \bf Teacher ID  & \bf \#1 & \bf \#2 & \bf \#3  & \bf \#4 & \bf \#5 & \bf \#6  & \bf \#7  & \bf \#8  & \bf \#9  & \bf \#10 \\
\midrule
\multirow{2}{*}{\bf CIFAR-10} & Transfer Acc. (\%) & 62.13 & 61.38 & 63.73 & \cellcolor{gray!50} 66.51 & 63.28  & 65.48 & 62.41  & 62.97 & 63.93  & 61.99  \\
& Saliency Sim.  & 0.35  & 0.61 & 0.92 & \cellcolor{gray!50} 1.0 & 0.11 & 0.61 & 0.0 & 0.26  & 0.53 & 0.19 \\
\midrule
\multirow{2}{*}{\bf CIFAR-100} & Transfer Acc. (\%) & 31.76 & 33.17 & 30.19 & 39.04 &  41.04  & 42.17 & 41.53  & 41.82 & \cellcolor{gray!50} 43.89  & 39.27 \\
& Saliency Sim.  & 0.43  & 0.23 & 0.0 & 0.25 & 0.42 & 0.74 & 0.48 & 0.42  & \cellcolor{gray!50} 1.0 & 0.77 \\
\bottomrule
\end{tabular}
}
\caption{Relation between transfer accuracy and saliency similarity. The similarity is normalized to $[0, 1]$ for contrast. High saliency typically reveals better transfer performance, which validates the effectiveness of saliency-based knowledge selection.}
\label{tab:transfer}
\end{table*}

\paragraph{\textbf{Visualization of Dynamic Knowledge Selection.}}
With limited annotated data, the target model cannot have a great insight into the target model. The knowledge of the target model increases gradually with the training process. In this part, the saliency map of different stages when training on CIFAR10 and CIFAR100 is visualized in Figure \ref{fig:vis_trans}. It indicates that the most proper models for the target model are changed, especially from the first knowledge selection to the second. The visualization indicates that the knowledge selection needs to be dynamic. It is more in line with the characteristics of progressive learning.

\paragraph{\textbf{Analytical Experiments.}}
In federated learning, the number of models chosen to do the model aggregation may influence the performance of the global model. To further evaluate the effectiveness of our proposed method, we conduct several experiments on the different number of nodes selected from clients. Three nodes are selected from ten clients in single-task experiments to do the global model aggregation. In this section, the experiment settings are the same as single-task experiments settings, and the number of chosen nodes is from one to five. Results are shown in Figure \ref{fig:vis_trans}. From the results, selecting three nodes out of ten clients has desirable results and is moderately time-consuming. Fewer nodes may lead to limited knowledge transfer; more nodes may result in off-task knowledge transfer and unpredictable results. 

\paragraph{\textbf{The Effectiveness of Saliency Maps.}}
The saliency maps are used to measure the transferability in FedSA. Experiments are conducted to demonstrate the effectiveness of saliency maps in this section. First, we extract $10\%$ data from each class in CIFAR10 and CIFRA100 to construct probe datasets. 
Teacher models are numbered from 1 to 10, and all of them are pre-trained by local private datasets. Saliency map similarities are calculated with probe datasets between the global target model and the local teacher models respectively and then normalized to 0 to 1 for contrast. 
All parameters of the local teacher models, except the last layer, are frozen. Then the teacher models are trained to evaluate the transfer accuracy. From the results in Table \ref{tab:transfer}, the transfer accuracy and saliency similarity generally have a positive correlation. 

\begin{table}[t]
\renewcommand{\arraystretch}{0.9}
\centering
\small
\begin{tabular}{c | l c }
\toprule
\bf Task / Metric & \bf Method & \bf Performance \\ 
\midrule
\multirow{6}{*}{Cls. / Acc(\%)} & Scratch Training  & 31.47  \\
& Transfer  &  35.68 \\
& Pseudo Label\cite{lee2013pseudo}  & 26.60 \\
& Noisy Student\cite{xie2020self}  & 25.75  \\
& FedAvg$+$\cite{mcmahan2017communication} & 31.94 \\
& \bf FedSA & \bf 51.58  \\
\midrule
\multirow{6}{*}{Seg. / mIoU} & Scratch Training  & 0.07 \\
& Transfer  & 0.21  \\
& Pseudo Label\cite{lee2013pseudo}   & 0.08 \\
& Noisy Student\cite{xie2020self}  & 0.06 \\
& FedAvg$+$\cite{mcmahan2017communication} & 0.14 \\
& \bf FedSA & \bf 0.41  \\
\bottomrule
\end{tabular}
\caption{Performance of student model amalgamated from Taskonomy models. The classification and semantic segmentation are used as the target task for comparison.}
\label{tab:taskonomy}
\end{table}

\subsection{Multi-Task Amalgamation} 

\paragraph{\textbf{Results on Taskonomy.}}
To show the effectiveness of the proposed method on the cross-task settings, we conduct experiments on Taskonomy dataset~\cite{zamir2018taskonomy}.
Taskonomy provides 26 types of annotations for different visual tasks and the corresponding pre-trained models. In this section, we deploy FedSA to amalgamate knowledge from different models to resolve a different task. 

In taskonomy\cite{zamir2018taskonomy} settings, the original data is split into the training, validation, and test set. The training set is used as the original local private data for pre-trained models in this experiment. The validation set with annotations is used as the target data for the target task. The test set is used to test the performance of the method. One of the tasks in taskonomy is selected as the target task, while others are regarded as the pre-trained teacher models. In detail, the tasks, semantic segmentation, and object classification are selected as the target task to validate the effectiveness of the proposed FedSA.
Other 24 visual tasks also can be selected as the target task, and FedSA is suitable for all those tasks.
The results in Table \ref{tab:taskonomy} indicate the proposed FedSA is quite competitive compared with others.

\paragraph{\textbf{Visualization of Semantic Segmentation.}}
In this section, the semantic segmentation results of different methods are visualized in Figure \ref{fig:vis_seg} to show the effect of the proposed FedSA more intuitively. We choose four images from the test set in taskonomy\cite{zamir2018taskonomy} as the input of the target model and compare the proposed method FedSA with other baselines. In Figure \ref{fig:vis_seg}, the results of Scratch Training are unsatisfactory, for the amount of target data is limited. The results of Pseudo Label\cite{lee2013pseudo} and Noisy Student\cite{xie2020self} are also undesirable because the model cannot learn the knowledge from pre-trained models. Owing to the redundancy of knowledge, FedAvg$+$ does not have excellent performance. As for the Transfer method, the result gets improved effectively with the prior knowledge from the pre-trained models. In the proposed method FedSA, the appropriate knowledge from different models is amalgamated into the target model with task adaption. Therefore, the proposed method FedSA can predict a more accurate segmentation boundary than other methods.


\section{Discussion}
Overall, we propose an innovative method to solve model and data privacy protection and limited target data problem.
As far as we know, this problem is valuable, and no method directly addresses this problem. 
The method introduces selective aggregation to avoid the exposure of the pre-trained models. Although the knowledge distillation still risks privacy disclosure, the amalgamation of different local student models fuses the knowledge from different pre-trained models. It further guarantees the model privacy. In the single-task amalgamation experiment, all pre-trained models are for classification. The pre-trained models in multi-task amalgamation include various visual tasks. Although we only verify the method on classification and semantic segmentation, the method can be easily adapted to other visual tasks. 

\section{Conclusions}

In this work, we propose a new model reusing
scheme termed FedSA to learn a student model from private pre-trained models. 
Specifically, we devise an efficient 
saliency-based approach to estimate the importance 
of different teachers and selectively amalgamate useful knowledge from teachers.
Extensive experiments demonstrate that FedSA achieves performance 
superior to baseline methods across various datasets. 
In our future work, we will take a further step
towards layer-level federated selective aggregation even to strengthen knowledge transfer.


\clearpage
%
%
\bibliographystyle{splncs04}
\bibliography{egbib}
\end{document}